\documentclass[runningheads]{llncs}
\usepackage[T1]{fontenc}
\usepackage{graphicx}
\usepackage{booktabs}
\usepackage[misc]{ifsym}
\newcommand{\corr}{(\Letter)}
\usepackage{amsmath,amsfonts,bm}
\usepackage{algorithm}
\usepackage{algpseudocode}

\usepackage{amsmath,amsfonts,bm}

\def\1{\bm{1}}

\def\eps{{\epsilon}}

\def\vg{{\bm{g}}}

\def\vx{{\bm{x}}}

\DeclareMathAlphabet{\mathsfit}{\encodingdefault}{\sfdefault}{m}{sl}
\SetMathAlphabet{\mathsfit}{bold}{\encodingdefault}{\sfdefault}{bx}{n}

\def\gX{{\mathcal{X}}}
\def\gY{{\mathcal{Y}}}

\newcommand{\pdata}{p_{\rm{data}}}

\usepackage{xcolor}
\usepackage{import}
\usepackage{pifont}
\usepackage{subcaption} 
\usepackage{array} 

\newcolumntype{H}{>{\setbox0=\hbox\bgroup}c<{\egroup}@{}} 

\usepackage[textsize=tiny]{todonotes}

\newcommand{\pcl}{p_{\rm{cl}}}
\newcommand{\pt}{p_{\boldsymbol{\theta}}}

\newcommand{\xzeropred}{\hat{\vx}_0^{(\vx_t)}}
\newcommand{\xzeropredmone}{\hat{\vx}_0^{(\vx_{t-1})}}
\newcommand{\xzeroprediction}{\hat{\vx}_0^{(\vx_t)}\text{-prediction}}

\newcommand{\normaldist}{\mathcal{N}(0, I)}

\newcommand{\uniformdistdiscrete}[2]{\mathcal{U}\{#1, #2\}}

\def\pt{p_\theta}

\usepackage{hyperref}
\begin{document}

\title{Diffusion Classifier Guidance for Non-robust Classifiers}
\tocauthor{Philipp Vaeth, Dibyanshu Kumar, Benjamin Paassen, Magda Gregorová} 
\toctitle{Diffusion Classifier Guidance for Non-robust Classifiers}


\author{
Philipp Vaeth\inst{1,2}\orcidID{0000-0002-8247-7907}\corr
\and \\
Dibyanshu Kumar\inst{1}\orcidID{0009-0007-2542-4781} 
\and \\
Benjamin~Paassen \inst{2}\orcidID{0000-0002-3899-2450}
\and \\
Magda Gregorová\inst{1}\orcidID{0000-0002-1285-8130}}

\authorrunning{Vaeth et al.}
\institute{Center for Artificial Intelligence and Robotics, Technical University of Applied Sciences Würzburg-Schweinfurt, Franz-Horn-Straße 2,
Würzburg, Germany\\
\email{\{philipp.vaeth,magda.gregorova\}@thws.de, kumardibyanshu05@gmail.com} \and 
Bielefeld University, Universitätsstraße 25, Bielefeld, Germany\\
\email{bpaassen@techfak.uni-bielefeld.de}}

\maketitle              

\begin{abstract}
Classifier guidance is intended to steer a diffusion process such that a given classifier reliably recognizes the generated data point as a certain class. 
However, most classifier guidance approaches are restricted to robust classifiers, which were specifically trained on the noise of the diffusion forward process. 
We extend classifier guidance to work with general, non-robust, classifiers that were trained without noise.
We analyze the sensitivity of both non-robust and robust classifiers to noise of the diffusion process on the standard CelebA data set, the specialized SportBalls data set and the high-dimensional real-world CelebA-HQ data set. 
Our findings reveal that non-robust classifiers exhibit significant accuracy degradation under noisy conditions, leading to unstable guidance gradients. 
To mitigate these issues, we propose a method that utilizes one-step denoised image predictions and implements stabilization techniques inspired by stochastic optimization methods, such as exponential moving averages. 
Experimental results demonstrate that our approach improves the stability of classifier guidance while maintaining sample diversity and visual quality. 
This work contributes to advancing conditional sampling techniques in generative models, enabling a broader range of classifiers to be used as guidance classifiers.

\\\\
\textbf{Reproducibility:} The code, the trained model weights and the supplementary material to reproduce the results is available at \url{https://github.com/philippvaeth/nrCG}.

\end{abstract}

\newpage

\section{Introduction}\label{sec:introduction}
Denoising diffusion probabilistic models (DDPM)~\cite{ddpm} are state of the art generative models, modelling an intractable data distribution $\vx_0 \sim \pdata$ via a learned latent variable model $\pt(\vx_0) = \int\, p(\vx_T)\, \prod_{t=1}^T \pt(\vx_{t-1} \mid \vx_t)\, d \vx_{1:T}$. 
Through a Markov chain Gaussian forward process ($\vx_0 \rightarrow \vx_T$) with noising transitions ${q(\vx_t \mid \vx_{t-1}) :=\mathcal{N}\left(\vx_t ; \sqrt{1-\beta_t} \vx_{t-1}, \beta_t \mathbf{I}\right)}$, the data $\vx_0$ is progressively noised with a pre-defined variance schedule $\beta_1, \ldots, \beta_T$.
The Gaussian Markov reverse process ($\vx_T \rightarrow \vx_0$) with learned denoising steps $p_\theta\left(\vx_{t-1} \mid \vx_t\right)$ reverses the forward process from random noise $\vx_T \sim \normaldist$ to produce samples following the data distribution $\pt \approx \pdata$.

A special property of this type of generative model is the iterative sampling procedure where conditional information can be added without the need for training a specific conditional model through a procedure known as \textit{classifier guidance}~\cite{diffusion_thermo,dhariwal2021diffusion}.
For an unconditionally trained DDPM $\pt(\vx_{t-1} \mid \vx_t) = \mathcal{N}(\vx_{t-1}; \mu_\theta(\vx_t), \Sigma_t(\vx_t))$, the mean $\mu_\theta(\vx_t)$ of the transitions can be shifted by the gradients of a classifier trained over the noisy data $\vx_t$ as:
\begin{equation}
    \mu_\theta(\vx_t)' = \mu_\theta(\vx_t)
    + s \, \Sigma_t(\vx_t) \, \nabla_{\vx_t} \log \pcl\left(y \mid \vx_t\right)\enspace ,
    \label{eq:ClassifierGuidance}
\end{equation}
where $s$ is a gradient scaling factor controlling the strength of the classifier guidance, and $\mu_\theta(\vx_t)'$ is the new mean of the reverse transition used for conditionally sampling the previous sample $\vx_{t-1}$. 

Classifier guidance is commonly used to add conditional information during the diffusion reverse process (e.g., in explainability~\cite{dvce}, in protein design~\cite{gruver2024protein} and in molecular design~\cite{weiss2023guided}).
The main limitation of classifier guidance is that the classifier needs to be robust to noise similar to that added during the diffusion forward process~\cite{dhariwal2021diffusion} so that the gradients $\nabla_{\vx_t} \log \pcl\left(y \mid \vx_t\right)$ in equation~\ref{eq:ClassifierGuidance} are meaningful.
This requires training a guidance classifier for each specific diffusion model and re-training it if the desired conditioning changes or if the diffusion forward process definition changes.
Extending classifier guidance to classifiers not trained over the specific DDPM noise (non-robust classifiers) remains a challenge. 

A previously proposed solution is to let the classifier decide on a one-step denoised image from the diffusion model instead of the noisy images directly, referred to as $\xzeroprediction$~\cite{avrahami2022blended,dvce,gradcheck}.
We introduce the $\xzeroprediction$ in detail in section~\ref{sec:xzeropred} (equation~\ref{eq:xzeropred}), and show that it is not enough to solve the challenge of non-robust classifier guidance.
Based on a detailed analysis of the classifier gradients including the $\xzeroprediction$, we propose in section~\ref{sec:stabilization} to leverage methods from stochastic optimization to additionally stabilize the non-robust guidance gradients further, bridging the gap to the performance of robust classifier guidance.
Finally, we transfer our proposed stabilization method to the diffusion reverse process in section~\ref{sec:reverse} and show that the stabilization enables the use of non-robust classifiers for guided sampling.
In summary, we provide a detailed analysis of how non-robust and robust classifiers behave during the diffusion forward process, and propose a guidance stabilization technique that allows non-robust classifiers to be used effectively for guidance in the diffusion reverse process.

\section{Diffusion forward process}\label{sec:forward}
We start our analysis by comparing the classifier accuracy over different levels of noisy data in section~\ref{sec:acc}.
We then showcase how the logits~(section \ref{sec:logits}) and gradients~(section \ref{sec:grads}) of the classifier behave over time $t$ for similar inputs.
Finally in section~\ref{sec:stabilization}, we analyze how the $\xzeroprediction$ (equation~\ref{eq:xzeropred}) influences the gradients of non-robust classifiers and, as a result, propose stabilization techniques to further improve non-robust classifier guidance.
We conclude section~\ref{sec:forward} by recommending a stabilization technique for non-robust classifier guidance and test this on the reverse diffusion process in section~\ref{sec:reverse}.

For our analysis, we train two standard MobileNetV3~\cite{mobilenetv3} classifiers on the CelebA~\cite{celeba} data set with an image size of 64x64 (details in section~\ref{sec:reverse}) to detect the binary attribute \textit{female}: (1) a \textbf{non-robust} classifier trained on the original non-noisy data and (2) a \textbf{robust classifier} trained on data augmented by the forward noising process of the diffusion model.
In detail, for a standard training batch of $n$ images, we draw $n$ time steps from a discrete uniform distribution $t \sim \uniformdistdiscrete{0}{T}$ and run the diffusion forward process for each image as: 
\begin{equation}
    q\left(\vx_t \mid \vx_0\right)=\mathcal{N}\left(\vx_t ; \sqrt{\bar{\alpha}_t} \vx_0,\left(1-\bar{\alpha}_t\right) \mathbf{I}\right)\enspace,
    \label{eq:qxtgx0}
\end{equation}
with $\bar{\alpha}_t:=\prod_{s=1}^t \alpha_s$ and $\alpha_t:=1-\beta_t$~\cite{ddpm}.
An increasing $\beta_t$ noise schedule therefore corresponds to progressively noisier samples $\vx_t$ for a higher $t$.

For our diffusion model, we train a standard DDPM~\cite{ddpm} with a linear noise schedule ($\beta_0=0.0001, \beta_T=0.02$), $T=400$ diffusion steps, a standard U-Net architecture~\cite{unet} for the noise predictor, and the simplified MSE noise prediction training objective~\cite{ddpm}. 
Our diffusion models are implemented using the open-source Diffusers toolbox~\cite{von-platen-etal-2022-diffusers} and trained for 1000 epochs (around 3 days on a single NVIDIA A80 GPU).

\subsection{Accuracy of the classifiers on noisy data} \label{sec:acc}
\begin{figure}[!h]
    \centering
    \includegraphics[width=1\linewidth]{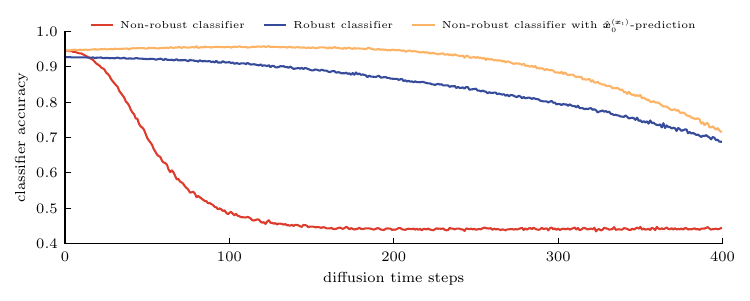}
    \caption{Classification accuracy comparison of the robust, non-robust, and non-robust with $\xzeroprediction$ (eq.~\ref{eq:xzeropred}) classifiers on the CelebA binary attribute female. The metric is reported as the average over the validation data set.}\label{fig:accuracy}
\end{figure}

In figure~\ref{fig:accuracy}, we compare the classification accuracy of the robust and the non-robust classifiers over the noisy validation data set (by applying equation~\ref{eq:qxtgx0}) and see that the non-robust classifier accuracy (red) drops significantly with increasing noise levels added through increasing diffusion steps, up to the point of random guessing at less than 25\% of the total diffusion steps $T$. 
This analysis of the classification performance is a simple way to understand classifier robustness over different noise levels.
However, the classification performance analysis works over the validation set perturbed by different amount of random noise, disregarding previous time steps (equation~\ref{eq:qxtgx0}).
In the diffusion forward process, the dependency on the previous sample is critical for the model definition and the sampling procedure (Markov property).
\newpage
\subsection{Sensitivity of the classifier logits} \label{sec:logits}
To further investigate the implications of low classification accuracy in the presence of noisy data points, we propose to analyze the sensitivity of the classifier's output scores (logits) to small changes in input features over time. 
This approach allows us to measure how the quantity of noise in the data points affects the decision boundary and robustness of the classifier.
Specifically, starting from the same image $\vx_0$, we do not sample two adjacent noisy versions $\vx_t$ and $\vx_{t-1}$ independently (equation~\ref{eq:qxtgx0}), but instead use the same noise to produce both noisy images.
This results in small changes by construction, where the same features are perturbed in $\vx_t$ and $\vx_{t-1}$ but at different scales based on the $\beta$ schedule of the DDPM forward process.
This is in line with the diffusion forward process definition in section~\ref{sec:introduction}, in which $\vx_t$ is a more noisy version of $\vx_{t-1}$.
We consider a classification function $f : \gX
 \to \gY^{D}$ which maps an RGB input image 
 to a $D$-dimensional vector of class logits and define the metric $S_l$ as:
\begin{equation}
    \begin{split}
        S_{l}(\vx_t,\vx_{t-1}) &=  \frac{\|\,f(\vx_t)-\,f(\vx_{t-1})\|_2}{\|\vx_t-\vx_{t-1}\|_2} \enspace.  \\
    \end{split} 
    \label{eq:Sl}
\end{equation}

\begin{figure}[!ht]
    \centering
    \includegraphics[width=1\linewidth]{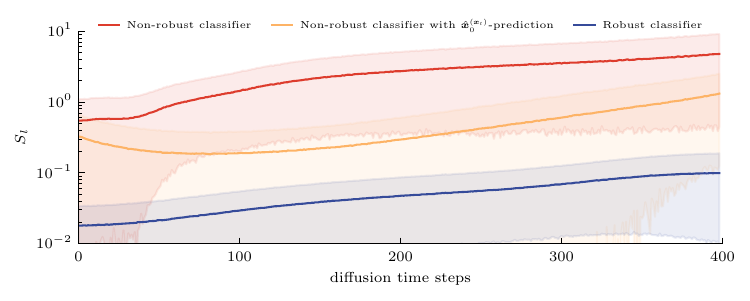}
    \caption{Logit sensitivity $S_l$ (log scale) as defined in eq.~\ref{eq:Sl} over time $t$ for the robust, non-robust, and non-robust with $\xzeroprediction$ (eq.~\ref{eq:xzeropred}) classifiers on CelebA. The metric is reported as the average (and std) over the validation data set.}
    \label{fig:logits}
\end{figure}

For two noisy data points $\vx_t$ and $\vx_{t-1}$ on the same diffusion trajectory (starting from the same $\vx_0$), small differences between these points should correspond to small differences in logits for a robust classifier. 
Note that the metric $S_l$ is similar to the discrete approximation of derivatives.
We compare the score $S_l$ (equation~\ref{eq:Sl}) over the entire diffusion forward process for our classifiers in figure~\ref{fig:logits} to analyze the noise sensitivity of classifier logits over time.
The results confirm that the non-robust classifier is indeed much more sensitive to small input changes than the robust classifier.
This means that the non-robust classifier function is not smooth and reacts with different output logits for small input perturbations, hinting to possibly undesired behavior for the guidance of the diffusion reverse process based on classifier gradients. 
\subsection{Stability of the classifier gradients} \label{sec:grads}
Going a step further beyond logits, we can directly compute gradients just as they would be used in the sampling process of the diffusion model to confirm that unstable logits over time $t$ indeed affect the gradients necessary in classifier guidance.
We run the same experiment, but compare the sensitivity of gradients over time $t$ instead of logits:
\begin{equation}
    \begin{split}
        S_{g}(\vx_t,\vx_{t-1}) &=  \frac{\|\,\nabla_{\vx_t} f(\vx_t)-\,\nabla_{\vx_{t-1}} f(\vx_{t-1})\|_2}{\|\vx_t-\vx_{t-1}\|_2} \enspace.  \\
    \end{split} 
    \label{eq:Sg}
\end{equation}
An alternative interpretation of equation~\ref{eq:Sg} is in terms of geometry. 
Equation~\ref{eq:Sg} quantifies to what degree the guidance vectors point in similar directions for adjacent time steps $t$ and $t-1$. 
We note that $S_g$ is connected to the discrete approximation of second-order derivatives, that is the curvature of the classification function over time.
In practice, a low $S_g$ score would correspond to gradual introduction of features during conditional diffusion sampling instead of sudden feature changes.
Based on this intuition, we can quantify through the metric $S_g$ how informative classifier gradients are for conditional sampling.

\begin{figure}[!ht]
    \centering
    \includegraphics[width=1\linewidth]{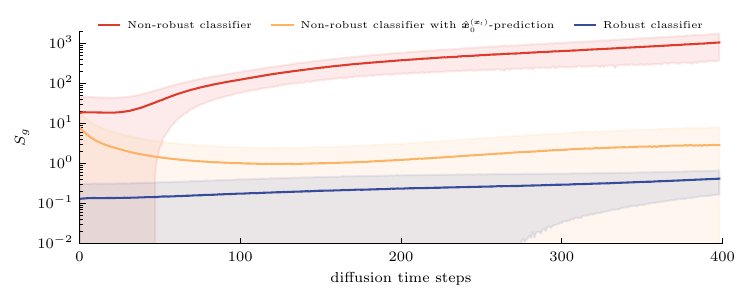}
    \caption{Gradient sensitivity $S_g$ (log scale) as defined in eq.~\ref{eq:Sg} over time $t$ for the robust, non-robust, and non-robust with $\xzeroprediction$ (eq.~\ref{eq:xzeropred}) classifiers on CelebA. The metric is reported as the average (and std) over the validation set.}\label{fig:grads}
\end{figure}

In figure~\ref{fig:grads}, we show the metric $S_g$ over time for the same experimental setup as previously, confirming that the non-robust classifier with unstable logit outputs (as demonstrated in figure~\ref{fig:logits}), indeed does not have informative gradients and is therefore not suitable for conditional guidance.
On the contrary, the robust classifier (blue line in figure~\ref{fig:grads}) shows low gradient sensitivity as measured by $S_g$, enabling the use of the robust classifier for classifier guidance.
\newpage
\subsection{Informative classifier gradients through $\xzeroprediction$} \label{sec:xzeropred}
To summarize, we have shown that classifying noisy data points with a classifier not trained over the same noise results in a loss of accuracy and high sensitivity of the classifier outputs to such noise.
We have also shown that this results in unstable and therefore non-informative gradients, which are not suitable for the use in classifier guidance.
One approach to resolve this issue is to apply the classifier not on the noisy diffusion data $\vx_t$ but on an approximation of the fully denoised image $\vx_0$~\cite{avrahami2022blended,dvce,gradcheck}. 
We can estimate the $\xzeroprediction$ via:
\begin{equation}
    \xzeropred = \frac{\vx_t}{\sqrt{\bar{\alpha}_t}} - \frac{\sqrt{1 - \bar{\alpha}_t}}{\sqrt{\bar{\alpha}_t}}\, \epsilon_\theta(\vx_t,t)\enspace.    
    \label{eq:xzeropred}
\end{equation}
The $\xzeroprediction$ seemingly resolves the issue of non-robust classifiers for classifier guidance, supported by the high classification accuracy over noisy data when applying the $\xzeroprediction$ before the classification (orange line in figure~\ref{fig:accuracy}).
In addition to classification accuracy, however, we also consider the gradient sensitivity for $\xzeroprediction$:
\begin{equation}
    \begin{split}
        \hat{S}_{g}(\vx_t,\vx_{t-1}) &=  \frac{\|\,\nabla_{\vx_t} f(\xzeropred)-\,\nabla_{\vx_{t-1}} f(\xzeropredmone)\|_2}{\|\vx_t-\vx_{t-1}\|_2} \enspace.  \\
    \end{split} 
    \label{eq:Sgx0}
\end{equation}

We show in figure~\ref{fig:grads}, that the classifier with $\xzeroprediction$ (orange line) substantially reduces gradient sensitivity (and hence improves gradient stability), but does not yet achieve the level of a robust classifier. 
Hence, we believe that further improvements, beyond the $\xzeroprediction$, are required.
We note that $\xzeroprediction$ dramatically increases memory cost because gradients need to be propagated not only through the classifier but also through the diffusion model at each denoising step.
 
\subsection{Stable classifier gradients through moving averages} \label{sec:stabilization}
We begin our improvements with the insight that the classifier guidance process in equation~\ref{eq:ClassifierGuidance} effectively acts as a moving average because the mean of the reverse sampling process in every step is the sum of the mean of the previous step and the classifier guidance vector. 
However, the guidance vectors are computed independently in each step $t$, meaning that their directions can drastically change between time steps as discussed in section~\ref{sec:grads} and shown in figures~\ref{fig:logits} and~\ref{fig:grads}. 
Accordingly, it stands to reason to adjust classifier guidance to explicitly perform a moving average over the guidance vectors, thus enhancing the gradient stability.
We explore two stabilization techniques inspired by the two most common stochastic optimization algorithms, SGD with momentum~\cite{sgdmomentum} and ADAM~\cite{adam}.
For a given guidance gradient $\vg$, momentum strength $\beta$ and $\eps > 0$, we define:

\begin{align}
    \nu^{\text{ema}}_t(\vg, \beta) &=  \beta\, \nu^{\text{ema}}_{t-1} +(1-\beta)\, \vg\enspace, \label{eq:ema}\\
    \nu^{\text{adam}}_t(\vg) &= \frac{\nu^{\text{ema}}_t(\vg, \beta = 0.9)}{\sqrt{\nu^{\text{ema}}_t(\vg^2, \beta = 0.999)} + \eps} \enspace. \label{eq:adam}
\end{align}\\

We do not include any de-biasing terms into equations~\ref{eq:ema} and~\ref{eq:adam} to compensate for extremely noisy samples with barely any signal in the initial denoising steps ($\vx_T, \vx_{T-1}, \dots$), which results in unreliable gradients.
We therefore omit these de-biasing terms deliberately to bias the guidance terms toward zero.
In the reverse process, this will avoid adding unreliable conditioning information early in the sampling steps, which could potentially break the diffusion sampling process due to unlikely starting points.
\begin{figure}[!h]
    \centering
    \includegraphics[width=1\linewidth]{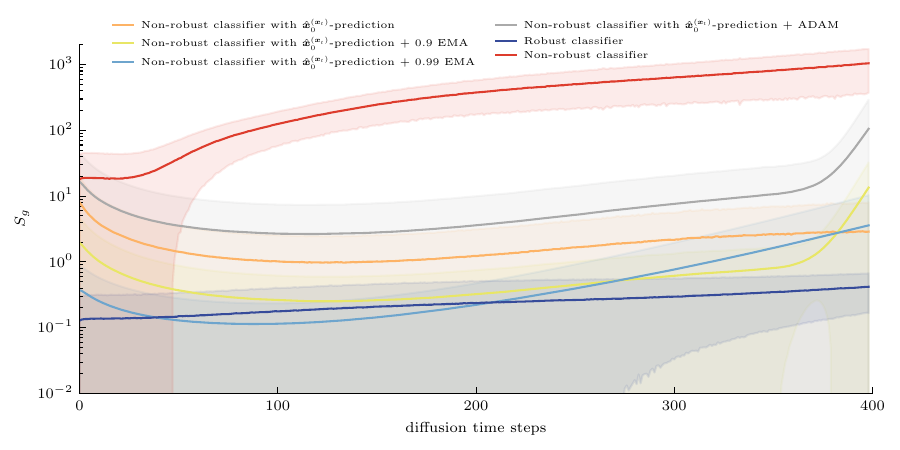}
    \caption{Gradient sensitivity $S_g$ (log scale) over time $t$ for the robust, non-robust, and non-robust with $\xzeroprediction$ (eq.~\ref{eq:xzeropred}) classifiers, as well as multiple stabilization techniques (eq.~\ref{eq:ema},\ref{eq:adam}). The metric is reported as the average (and std) over the CelebA validation data set.}\label{fig:grads_ema}
\end{figure}

We again experimentally validate on the forward process how these stabilization techniques change the gradient stability over time.
For this, we apply both techniques (equations~\ref{eq:ema}~and~\ref{eq:adam}) directly on the gradients in our gradient stability metric $S_g$ (equation~\ref{eq:Sgx0}). 
We show the gradient stability over time $t$ in figure~\ref{fig:grads_ema}, contrasting the robust classifier to the non-robust classifier with $\xzeroprediction$ and with the stabilization techniques.
For ADAM, the gradient stability deteriorates over increasing time $t$ due to the rescaling of the gradients by the running estimate of the second moment (see equation~\ref{eq:adam}), amplifying differences between time steps $t$ and $t-1$ based on the variance of the gradient (denominator of equation~\ref{eq:adam}).
For exponential moving averaging, the differences between neighboring diffusion time steps become naturally smaller, with a larger window size ($\beta=0.99$) contributing to even more stability over time.
Interestingly, the EMA stabilization with the large window size reaches the gradient stability of the robust classifier, especially during the first half of the forward process ($t < 200$).
\newpage
Our analysis demonstrates that gradient stability, measured by pairwise differences over time, is connected to the classifier accuracy. 
Additionally, we show that $\xzeroprediction$ enhances gradient quality from this perspective.
Furthermore, by explicitly enforcing stable feature changes over the diffusion time steps $t$ through exponential moving averaging of classifier gradients, we can bridge the gap between the non-robust classifier and the robust classifier in terms of gradient stability.
These observations have so far been on the diffusion forward process to observe the gradient behavior in isolation without interference of the diffusion reverse (sampling) process.
In section~\ref{sec:reverse}, we will translate these findings to the diffusion reverse process.

\section{Diffusion reverse process}\label{sec:reverse}

In the diffusion reverse process, we apply the techniques from section~\ref{sec:forward} to diffusion sampling.
In algorithm~\ref{alg:sampling}, we provide details about the implementation of our guided sampling setup.
The only difference to the standard DDPM classifier guidance are lines 3 and 4, which is where we apply our stabilization techniques.
We introduce the data sets in section~\ref{sec:datasets}, define the metrics used to evaluate the generated samples in section~\ref{sec:metrics}, and then use algorithm~\ref{alg:sampling} in section~\ref{sec:resultsanddiscussion} to produce conditional samples for our non-robust classifiers.

\begin{algorithm}[H]
\caption{Guided DDPM Sampling}\label{alg:sampling}
    \begin{algorithmic}[1]
    \State $\vx_T \sim \normaldist$, classifier guidance scale $s$, unconditionally trained DDPM $\mu_\theta(\vx_t)$, DDPM forward process variance $\Sigma_t(\vx_t)$, guidance stabilization function $\nu$
    \For{$t=T, \ldots, 1$}
        \State $\vg = \nabla_{\vx_t} \log \pcl\left(y \mid \xzeropred\right) \text { if } \xzeroprediction \text { (eq.\ref{eq:xzeropred}), else } \nabla_{\vx_t} \log \pcl\left(y \mid \vx_t\right)$
        \State $\vg = \nu(\vg) \text { if guidance-stabilization}$ \Comment{See eq.~\ref{eq:ema} and eq.~\ref{eq:adam}}
        \State $\vx_{t-1} = \mathcal{N}(\vx_{t-1}; \mu_\theta(\vx_t), \Sigma_t(\vx_t))$ \Comment{See diffusion reverse transition in sec.~\ref{sec:introduction}}
        \State $\vx_{t-1}' = \vx_{t-1} + s \, \Sigma_t(\vx_t) \, \nabla_{\vx_t}\, \vg$ \Comment{See eq.~\ref{eq:ClassifierGuidance}}
    \EndFor\\
    \Return $\vx_{0}'$
    \end{algorithmic}
\end{algorithm}

\subsection{Data sets}\label{sec:datasets}
For the data sets used in conditional sampling, we chose CelebA~\cite{celeba} as a standard image generation benchmark data set, use the SportBalls data set~\cite{xaidiff} as a custom data set specifically created for conditional generations, and use Celeba-HQ~\cite{celebahq} as the real-world high resolution data set with an off-the-shelve diffusion model.
We train a standard MobileNetV3~\cite{mobilenetv3} classifier on all data sets.
We train the non-robust classifier without data augmentation and the robust classifier with the training data corrupted by the same noise occurring in the diffusion forward process.

For the CelebA data set (64x64), we train the classifiers on the binary attribute \textit{female} (58.3\% of the total images).
This class was chosen based on easily distinguishable features of the classes and the relatively clear classification boundary.
The classifiers are trained on over more than 160k training images and evaluated over 20k validation images.

For a synthetic, more controllable, conditional sampling setup we use the SportBalls data set (64x64). 
The custom data set is created by randomly selecting one out of three sport balls (multi-class classification) and placing them at random coordinates on white background with random rotation and scaling.
The data set is carefully created to have similar objects (i.e., scaling, shape, size, rotation and placement) but with clear semantic differences (i.e., colors and pattern).
This data set is specifically constructed for conditional sampling due to clear class boundaries with balanced classes for the classifier and the white background for unambiguous generations without artifacts in the images.
The goal for the conditional DDPM sampling is to generate \textit{baseballs}.
The classifiers are trained on 80k training images and evaluated on 20k validation images.

For the real-world use-case on CelebA-HQ-256 (256x256), we train the same simple classifier just as for the other data sets, but use the pre-trained DDPM model from~\cite{ddpm}.
We use the DDPM model without modification in our stabilized guided sampling setup to showcase how our contribution translates to third-party models and higher dimensional data.
The class to generate is \textit{female} (64.1\% of the total images).
The non-robust classifier is trained on 28k training images and evaluated on 2k validation images.

\subsection{Metrics}\label{sec:metrics}
To evaluate the resulting samples on the CelebA and SportBalls data sets, we compute all following metrics over 50176 conditionally generated samples for the different stabilization setups (3-7 hours on a single NVIDIA A80 graphics card).
For Celeba-HQ, we compute the metrics over 1024 samples (4 hours on a single NVIDIA A80 graphics card) due to computational constraints.

To quantify if the guiding classifier successfully introduced class-conditional features, we apply the classifier on the final generated samples and compute the accuracy for the target class.
Different stabilization setups and guidance scales will lead to higher accuracy at the expense of image quality and diversity.

A common metric to quantify the visual quality of generated images is the Fréchet inception distance (FID)~\cite{fid}, which compares statistics of extracted features from a pre-trained network between the training data and generated images.
For this comparison, we randomly draw the same amount of generated samples from the training data as we generate.
A low FID score indicates visual similarity of the generated samples to the training data.
This metric also serves as a measure for diversity, as generated samples with only class-specific features are generally less close to the training set with a diverse set of features.

To complement the accuracy and the unconditional FID metric for visual quality, we compute a class-specific FID score which only operates on the data of the target class (cFID).
In practice, this means we compare the statistics of the conditionally generated samples not to that of the entire data set but only to training images of the target class.
A low cFID score ensures that the generated samples are visually close to the ground-truth images of the target class, ignoring potential features of other classes.

\subsection{DDPM sampling with stabilized non-robust classifiers}\label{sec:resultsanddiscussion}
\begin{figure}[!ht]
    \centering
    \includegraphics[width=1\linewidth]{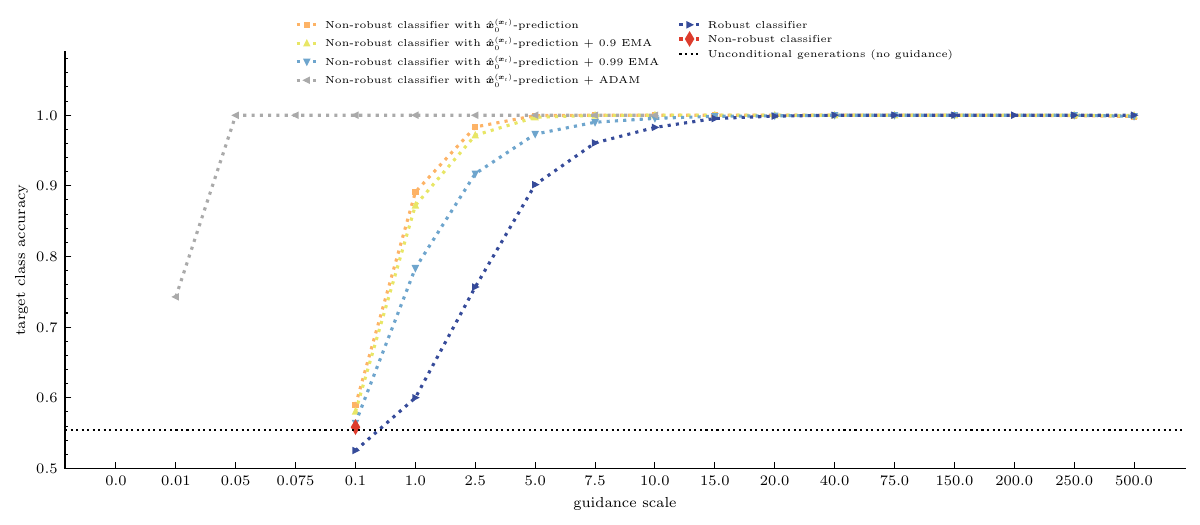}
    \caption{Accuracy comparison for conditional sampling on CelebA with various stabilization setups (eq.~\ref{eq:ema},\ref{eq:adam}). The accuracy is presented as the average over 50176 generated samples.}
    \label{fig:sampling_acc_celeba}
\end{figure}

\begin{figure}[t]
    \centering
    \includegraphics[width=1\linewidth]{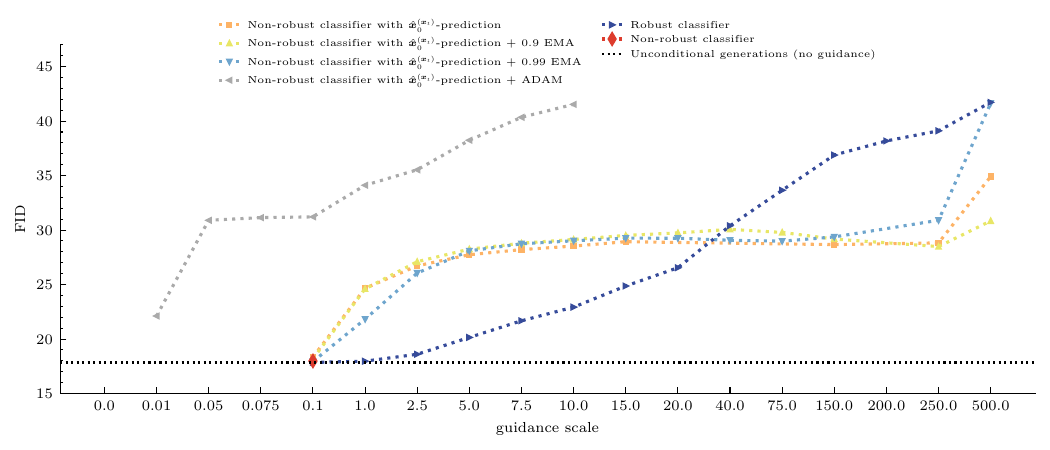}
    \caption{Unconditional FID comparison for conditional sampling on CelebA with various stabilization setups (eq.~\ref{eq:ema},\ref{eq:adam}). The unconditional FID is calculated over 50176 generated samples.}
    \label{fig:sampling_ufid_celeba}
\end{figure}

\begin{figure}[t]
    \centering
    \includegraphics[width=1\linewidth]{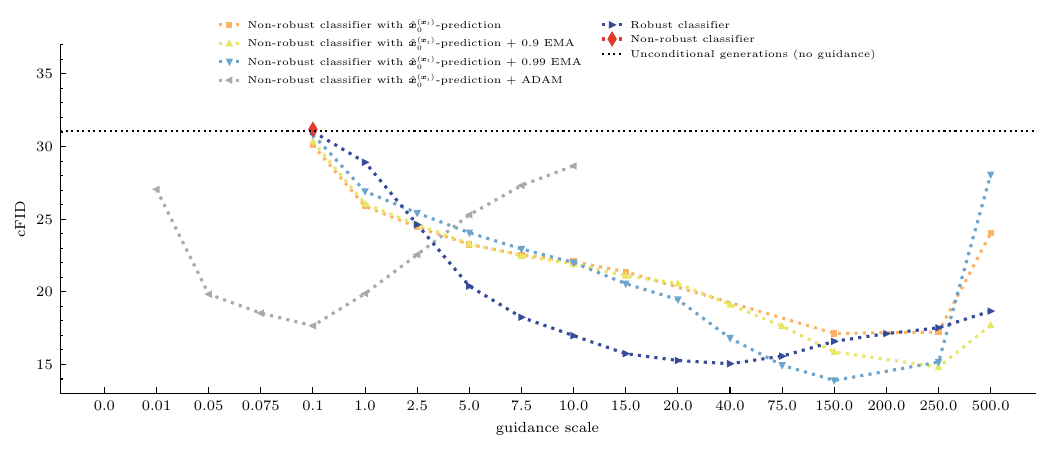}
    \caption{Target class FID comparison for conditional sampling on CelebA with various stabilization setups (eq.~\ref{eq:ema},\ref{eq:adam}). The class-conditional FID is calculated over 50176 generated samples.}
    \label{fig:sampling_cfid_celeba}
\end{figure}

We start our improved sampling experiments on the CelebA data set and conclude the section with experiments on the SportBalls and the CelebA-HQ data sets.
The key hyperparameter in classifier guidance is the scale $s$ (see algorithm~\ref{alg:sampling} and equation~\ref{eq:ClassifierGuidance}), known to trade-off class conditioning and sample diversity~\cite{dhariwal2021diffusion}.
We explore the robust classifier and the non-robust classifier with stabilization techniques, and present the accuracy over different guidance scales in figure~\ref{fig:sampling_acc_celeba}, the unconditional FID in figure~\ref{fig:sampling_ufid_celeba} and the conditional FID in figure~\ref{fig:sampling_cfid_celeba}.

We can observe that for a high enough guidance scale, all classifier setups except the non-robust classifier without stabilization techniques produce consistent class-conditional samples according to the classifier (figure~\ref{fig:sampling_acc_celeba}).
The non-robust classifier guidance fails without stabilization by offsetting the unconditional diffusion mean by so much that the diffusion reverse process can not recover, ultimately not producing any samples.
We can see that the ADAM stabilization requires a much lower scaling than the other stabilization techniques as the rescaling of the gradients by the variance (equation~\ref{eq:adam}) amplifies the guidance scale.

From the image quality as measured by the FID in figure~\ref{fig:sampling_ufid_celeba}, we can notice the FID of the robust classifier increases with a higher guidance scale.
This is expected behavior, as the quality of the images measured by the closeness to the entire data set should decrease if the diffusion model is constrained to generate features of one class only and therefore loses diversity.
The non-robust classifier with $\xzeroprediction$ and ADAM stabilization exhibits a more rapid increase in FID as the guidance scale increases, compared to the robust classifier.
The non-robust classifiers with $\xzeroprediction$ and with $\xzeroprediction$ + EMA stabilization all exhibit similar behavior with an increasing FID just as the target class accuracy increases. 
This levels off to a stable FID value even for very high guidance scales until too much guidance strength (>500 in this case) eventually increases the FID again. 
This indicates an optimal range just before this increase, where substantial guidance strength can be applied without compromising sample quality or overwhelming the diffusion process.

For the class-conditional FID in figure~\ref{fig:sampling_cfid_celeba}, we observe a decrease in cFID score with more guidance strength for the robust classifier up to a turning point (here $s>40$) when the cFID score increases again.
This means for higher guidance strength the overall sample quality (FID) decreases due to lower diversity as compared to the complete mixed-class training set, but the class-conditional sample quality (cFID) increases.
However, if the guidance strength is too high, the cFID reaches a turning point where the conditioning overpowers the diffusion process, generating samples not coherent with the underlying data distribution.
A similar behavior is shown by all guidance setups, highlighting that the choice of guidance strength trades-off sample quality and class conditioning.
Our proposed guidance setup, using $\xzeroprediction$ and EMA stabilization with $\beta=0.99$, achieves the best cFID score (13.9) while maintaining good overall image quality (FID of 29.37). 
This guidance setup outperforms even the unmodified robust classifier, demonstrating the potential of non-robust classifiers for conditional sampling when appropriately stabilized.
This confirms our findings from the diffusion forward process analysis in section~\ref{sec:forward}.
By introducing the $\xzeroprediction$ into the classifier guidance, the classifier gradients of the non-robust classifier are more meaningful.
Through exponential moving averaging of the classifier gradients, we can enforce stable feature changes over the guided reverse diffusion process.
The combination of the $\xzeroprediction$ and the exponential moving average of the gradients leads to successful classifier guidance even for the non-robust classifier.
We show generations for our best guidance setup as well as without guidance in table~\ref{tab:sampling_celeba}.
More images are provided in the supplementary material.

\begin{table}[h!]
    \centering  
    \caption{Metrics and first 10 samples for unconditional diffusion sampling (left) and conditional diffusion sampling (right) with the non-robust classifier, $\xzeroprediction$ (eq.~\ref{eq:xzeropred}), 0.99-EMA stabilization (eq.~\ref{eq:ema}) and guidance scale of 150.0 on CelebA. Metrics calculated over a batch of 50176 samples. More images are shown in the supplementary material.} \label{tab:sampling_celeba}  
    \begin{tabular}{c|c@{}}
        \toprule
        Unconditional & Conditional (best) \\ 
        \midrule
         \includegraphics[width=0.48\linewidth]{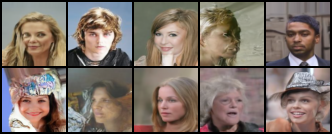} & \includegraphics[width=0.48\linewidth]{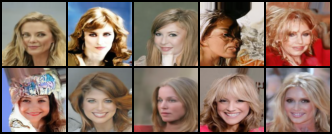} \\ \midrule
         {\textit{acc.:}} 55.67\%, {\textit{FID:}} 17.68 , {\textit{cFID:}} 30.96 & {\textit{acc.:}} 99.99\%, {\textit{FID:}} 29.37 , {\textit{cFID:}} 13.90 \\
        \bottomrule 
    \end{tabular}
\end{table}

\begin{figure}[!ht]
    \centering
    \begin{subfigure}[t]{0.49\textwidth}
        \includegraphics[width=1\linewidth]{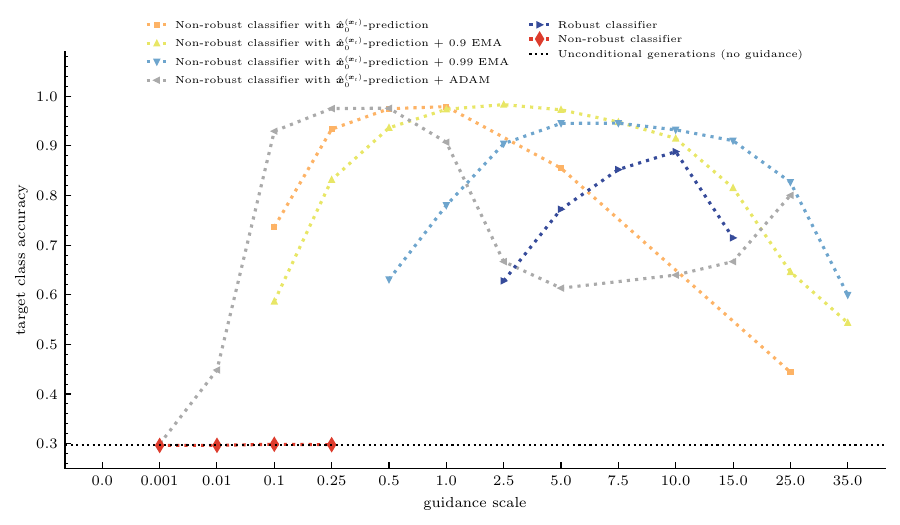}
        \caption{Accuracy}
        \label{fig:sampling_acc_sportballs}
    \end{subfigure}
    \begin{subfigure}[t]{0.49\textwidth}
        \includegraphics[width=1\linewidth]{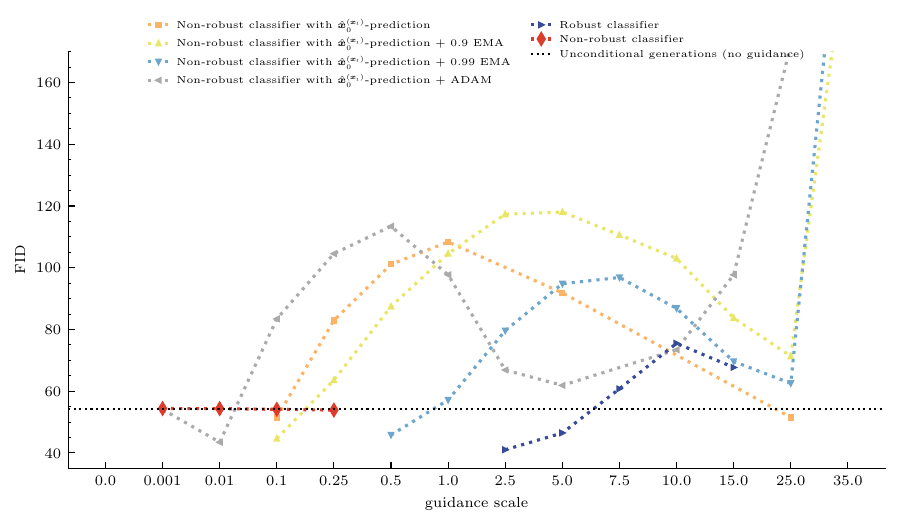}
        \caption{Unconditional FID}
        \label{fig:sampling_ufid_sportballs}
    \end{subfigure}
    \\
    \begin{subfigure}[t]{1\textwidth}
        \includegraphics[width=1\linewidth]{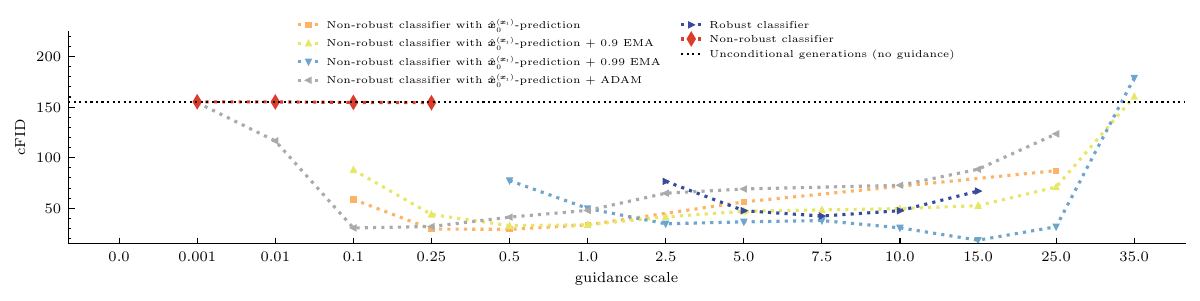}
        \caption{Conditional FID}
        \label{fig:sampling_cfid_sportballs}
    \end{subfigure}
    \caption{Accuracy, FID and cFID metrics for conditional sampling on SportBalls with stabilization setups (eq.~\ref{eq:ema},\ref{eq:adam}) calculated over 50176 generated samples.}
\end{figure}
We repeat the previous experiment on the more controlled multi-class SportBalls data set.
We present the accuracy in figure~\ref{fig:sampling_acc_sportballs}, the unconditional FID in figure~\ref{fig:sampling_ufid_sportballs} and the conditional FID in figure~\ref{fig:sampling_cfid_sportballs} for the different guidance setups and scaling strength.
Sample images are shown in figure~\ref{tab:sampling_sportballs}.
The robust classifier generates class-conditional samples, trading off the overall image quality with the amount of conditioning added to the diffusion reverse process (figure~\ref{fig:sampling_cfid_sportballs}).
All guidance setups improve the guidance mechanism for the non-robust classifier, with the 0.99-EMA stabilization reaching the lowest cFID score of 18.5 while maintaining good overall image quality with a FID of 69.6.
The guidance by the non-robust classifier fails similarly as on the CelebA data without any stabilization techniques.
The special setup of the data set with clear class boundaries and unambiguous class features is visible in the results, where the cFID decreases drastically when classifier guidance is successfully applied (table~\ref{tab:sampling_sportballs}).

\begin{table}[!ht]
    \centering  
    \caption{Metrics and first 10 samples for unconditional diffusion sampling (left) and conditional diffusion sampling (right) with the non-robust classifier, $\xzeroprediction$ (eq.~\ref{eq:xzeropred}), 0.99-EMA stabilization (eq.~\ref{eq:ema}) and guidance scale of 15.0 on the SportBalls data set. Metrics calculated over a batch of 50176 samples. More images are shown in the supplementary material.} \label{tab:sampling_sportballs}  
    \begin{tabular}{c|c@{}}
        \toprule
        Unconditional & Conditional (best) \\ 
        \midrule
         \includegraphics[width=0.49\linewidth]{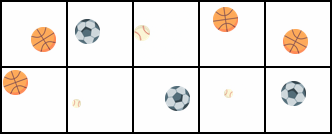} & \includegraphics[width=0.49\linewidth]{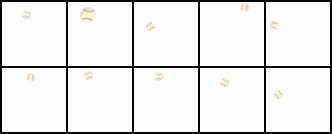} \\ \midrule
         {\textit{acc.:}} 29.63\%, {\textit{FID:}} 54.37 , {\textit{cFID:}} 155.19 & {\textit{acc.:}} 91.05\%, {\textit{FID:}} 69.64 , {\textit{cFID:}} 18.52 \\
        \bottomrule 
    \end{tabular}
\end{table}

\begin{table}[!ht]
    \centering  
    \caption{Metrics and samples for unconditional diffusion sampling (left) and conditional diffusion sampling (right) with $\xzeroprediction$ (eq.~\ref{eq:xzeropred}), 0.99-EMA stabilization (eq.~\ref{eq:ema}) and guidance scale of 10.0 on CelebA-HQ. Metrics calculated over a batch of 1024 samples. The top row shows the first 5 generations, the bottom row shows 5 hand-picked seeds for which the unconditional model produces the class male. More images are shown in the supplementary material.} \label{tab:sampling_celebahq}  
    \begin{tabular}{Hc|c@{}}
        \toprule
        & Unconditional & Conditional \\ 
        \midrule
        Metrics& \includegraphics[width=0.49\linewidth]{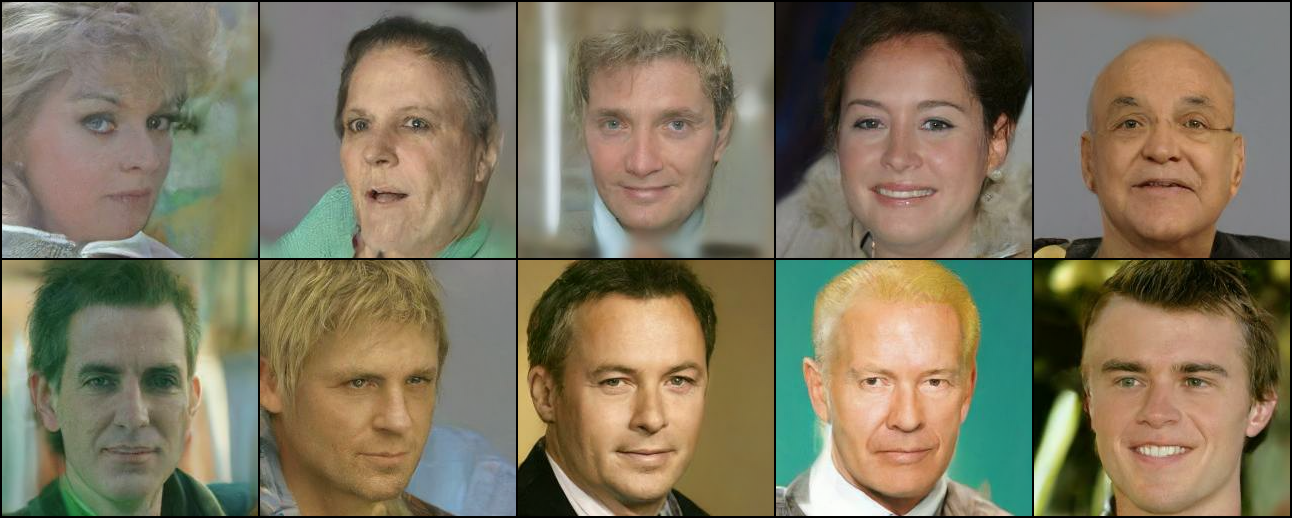} & \includegraphics[width=0.49\linewidth]{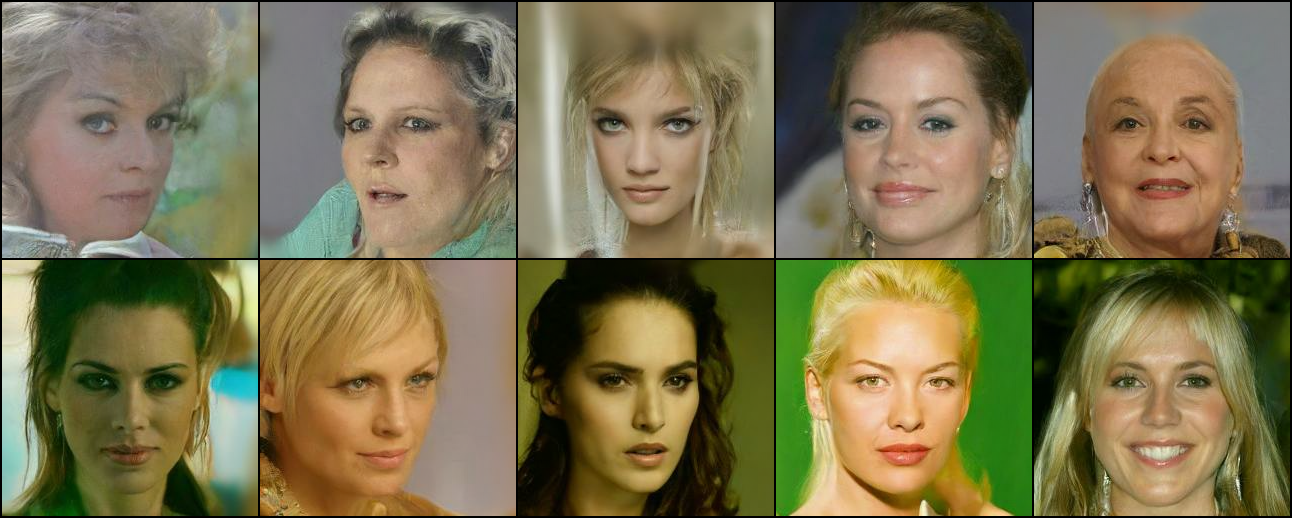} \\ \midrule
        Accuracy & {\textit{acc.:}} 77.1\%, {\textit{FID:}} 45.57 , {\textit{cFID:}} 49.70 & {\textit{acc.:}} 99.02\%, {\textit{FID:}} 50.67 , {\textit{cFID:}} 42.18 \\
        \bottomrule 
    \end{tabular}
\end{table}

For our real-world CelebA-HQ data set, we test our best guidance setup on an off-the-shelve DDPM.
We evaluate the metrics on 1024 samples and show the first 10 generated samples as well as the corresponding metrics in figure~\ref{tab:sampling_celebahq}.
The stabilized classifier guidance for the non-robust classifier with $\xzeroprediction$ and 0.99-EMA successfully generates class-conditional samples by achieving >~99\% target class accuracy, reducing the cFID by 7.52 points and trading-off conditioning with overall sample quality (FID increased by 5.1 points).
Visually, male faces are slightly altered towards what the classifier believes are female features.

\newpage
\section{Related Work}
Classifier guidance as proposed in~\cite{dhariwal2021diffusion} requires a classifier, which was trained on the same noise as introduced in the diffusion forward process.
A one-step estimate of the denoised image from the diffusion model was proposed to apply classifier guidance to noise-unaware classifiers~\cite{avrahami2022blended}, which we refer to as $\xzeroprediction$ in our paper. 
In combination with the $\xzeroprediction$, a robust classifier restricting the non-robust classifier gradients can be used in conjunction to enable guidance on arbitrary classifiers~\cite{dvce}.
However, to the best of our knowledge, no paper has so far specifically addressed the challenge of non-robust classifiers for classifier guidance without training a specialized classifier or diffusion model.
For classifier guidance with robust classifiers, multiple improvements have been suggested, for example~\cite{pixelasparam,progressiveguidance}.

Classifier-free guidance~\cite{classifierfreeguidance}, as the predecessor of classifier guidance, subsumes the auxiliary classifier into a Bayesian implicit classifier in the form of a conditional diffusion model.
Through training a conditional diffusion model, the unconditional and conditional denoising steps can be traded-off to achieve conditioning during sampling.
We mention this parallel line of work for completeness, but note that classifier-free guidance always requires training a conditional diffusion model, which therefore does not allow adding arbitrary conditioning information in the diffusion reverse process without retraining.

\newpage
\section{Conclusion}
In this study, we have extended classifier guidance techniques to non-robust classifiers within denoising diffusion probabilistic models (DDPMs). 
By addressing the inherent limitations of requiring specifically trained robust classifiers for classifier guidance, we built on top of previously proposed one-step denoised image predictions to stabilizes guidance gradients during the sampling process. 
Our findings demonstrate that incorporating stabilization techniques, particularly exponential moving averages, enhances gradient stability, bridging the performance gap between non-robust and robust classifiers.
The experimental results on the CelebA data set indicate that our approach not only improves classification accuracy but also maintains sample diversity and visual quality in generated images. 
Future work will focus on refining these methods and exploring their applicability to other generative models and diffusion samplers.
Especially other techniques from stochastic optimization and dynamic guidance schedules will be explored.

\textit{Limitations} Classifier-guidance is sensitive to hyperparameter choices, especially the guidance scaling.
We explored many hyperparameter choices in this study but did not specifically optimize for state-of-the-art FID scores.
We only explored two stabilization techniques based on SGD with momentum and ADAM as the two most commonly used methods from stochastic optimization.
This shows stabilization techniques are promising candidates to improve classifier guidance, other techniques not explored in this study may however improve gradient stability even further.
We also use the FID metric as-is with the feature extractor pre-trained on ImageNet~\cite{imagenet}. 
This results in higher FID values for CelebA and very high FID values for SportBalls, since the features in the data sets are different to features extracted on ImageNet.
In our analysis, we used one representative classifier architecture (MobileNetV3).
Other architectures may require different hyperparameter choices.
The same applies for the diffusion process, where we only used the standard DDPM setup.
Translating our findings to other diffusion reverse samplers is subject to future work.

\begin{credits}
\subsubsection{\ackname} This research is supported by the Center for Artificial Intelligence (CAIRO) at the Technical University of Applied Sciences Würzburg-Schweinfurt (THWS), Würzburg, Germany and the Bavarian Hightech Agenda.

\subsubsection{\discintname}
The authors have no competing interests to declare that are relevant to the content of this article.
\end{credits}

%
%
%
\bibliographystyle{splncs04}


\newpage
\begin{thebibliography}{10}
\providecommand{\url}[1]{\texttt{#1}}
\providecommand{\urlprefix}{URL }
\providecommand{\doi}[1]{https://doi.org/#1}

\bibitem{dvce}
Augustin, M., Boreiko, V., Croce, F., Hein, M.: Diffusion Visual Counterfactual Explanations. NeurIPS  (2022)

\bibitem{avrahami2022blended}
Avrahami, O., Lischinski, D., Fried, O.: Blended Diffusion for Text-driven Editing of Natural Images. CVPR  (2022)

\bibitem{imagenet}
Deng, J., Dong, W., Socher, R., Li, L.J., Li, K., Fei-Fei, L.: Imagenet: A Large-scale Hierarchical Image Database. CVPR  (2009)

\bibitem{dhariwal2021diffusion}
Dhariwal, P., Nichol, A.: Diffusion Models Beat Gans on Image Synthesis. NeurIPS  (2021)

\bibitem{pixelasparam}
Dinh, A.D., Liu, D., Xu, C.: Pixelasparam: A Gradient View on Diffusion Sampling with Guidance. ICML  (2023)

\bibitem{progressiveguidance}
Dinh, A.D., Liu, D., Xu, C.: Rethinking Conditional Diffusion Sampling with Progressive Guidance. NeurIPS  (2024)

\bibitem{gruver2024protein}
Gruver, N., Stanton, S., Frey, N., Rudner, T.G., Hotzel, I., Lafrance-Vanasse, J., Rajpal, A., Cho, K., Wilson, A.G.: Protein Design with Guided Discrete Diffusion. NeurIPS  (2024)

\bibitem{fid}
Heusel, M., Ramsauer, H., Unterthiner, T., Nessler, B., Hochreiter, S.: Gans Trained by a two Time-scale Update Rule Converge to a Local Nash Equilibrium. NeurIPS  (2017)

\bibitem{ddpm}
Ho, J., Jain, A., Abbeel, P.: Denoising Diffusion Probabilistic Models. NeurIPS  (2020)

\bibitem{classifierfreeguidance}
Ho, J., Salimans, T.: Classifier-free Diffusion Guidance. NeurIPS Workshop on Deep Generative Models and Downstream Applications  (2021)

\bibitem{mobilenetv3}
Howard, A., Sandler, M., Chu, G., Chen, L.C., Chen, B., Tan, M., Wang, W., Zhu, Y., Pang, R., Vasudevan, V., et~al.: Searching for Mobilenetv3. ICCV  (2019)

\bibitem{celebahq}
Karras, T., Aila, T., Laine, S., Lehtinen, J.: Progressive Growing of Gans for Improved Quality, Stability, and Variation. ICLR  (2018)

\bibitem{adam}
Kingma, D.P., Ba, J.: Adam: A Method for Stochastic Optimization. ICLR  (2015)

\bibitem{celeba}
Liu, Z., Luo, P., Wang, X., Tang, X.: Deep Learning Face Attributes in the Wild. ICCV  (2015)

\bibitem{von-platen-etal-2022-diffusers}
von Platen, P., Patil, S., Lozhkov, A., Cuenca, P., Lambert, N., Rasul, K., Davaadorj, M., Wolf, T.: Diffusers: State-of-the-art Diffusion Models. \url{https://github.com/huggingface/diffusers} (2022)

\bibitem{unet}
Ronneberger, O., Fischer, P., Brox, T.: U-net: Convolutional Networks for Biomedical Image Segmentation. MICCAI  (2015)

\bibitem{sgdmomentum}
Rumelhart, D.E., Hinton, G.E., Williams, R.J.: Learning Representations by Back-propagating Errors. Nature  (1986)

\bibitem{diffusion_thermo}
Sohl-Dickstein, J., Weiss, E., Maheswaranathan, N., Ganguli, S.: Deep Unsupervised Learning using Nonequilibrium Thermodynamics. ICML  (2015)

\bibitem{xaidiff}
Vaeth, P., Fruehwald, A.M., Paassen, B., Gregorova, M.: Generative Example-based Explanations: Bridging the Gap Between Generative Modeling and Explainability. arXiv:2410.20890  (2024)

\bibitem{gradcheck}
Vaeth, P., Fruehwald, A.M., Paassen, B., Gregorova, M.: Gradcheck: Analyzing Classifier Guidance Gradients for Conditional Diffusion Sampling. arXiv:2406.17399  (2024)

\bibitem{weiss2023guided}
Weiss, T., Mayo~Yanes, E., Chakraborty, S., Cosmo, L., Bronstein, A.M., Gershoni-Poranne, R.: Guided Diffusion for Inverse Molecular Design. Nature Computational Science  (2023)

\end{thebibliography}
%
\newpage
\appendix
\def \imgwidth {0.75}
\section{Supplementary material (compact version)}
\begin{figure}[!h]
\centering
\begin{subfigure}{.5\textwidth}
    \centering
    \includegraphics[width=\imgwidth\linewidth]{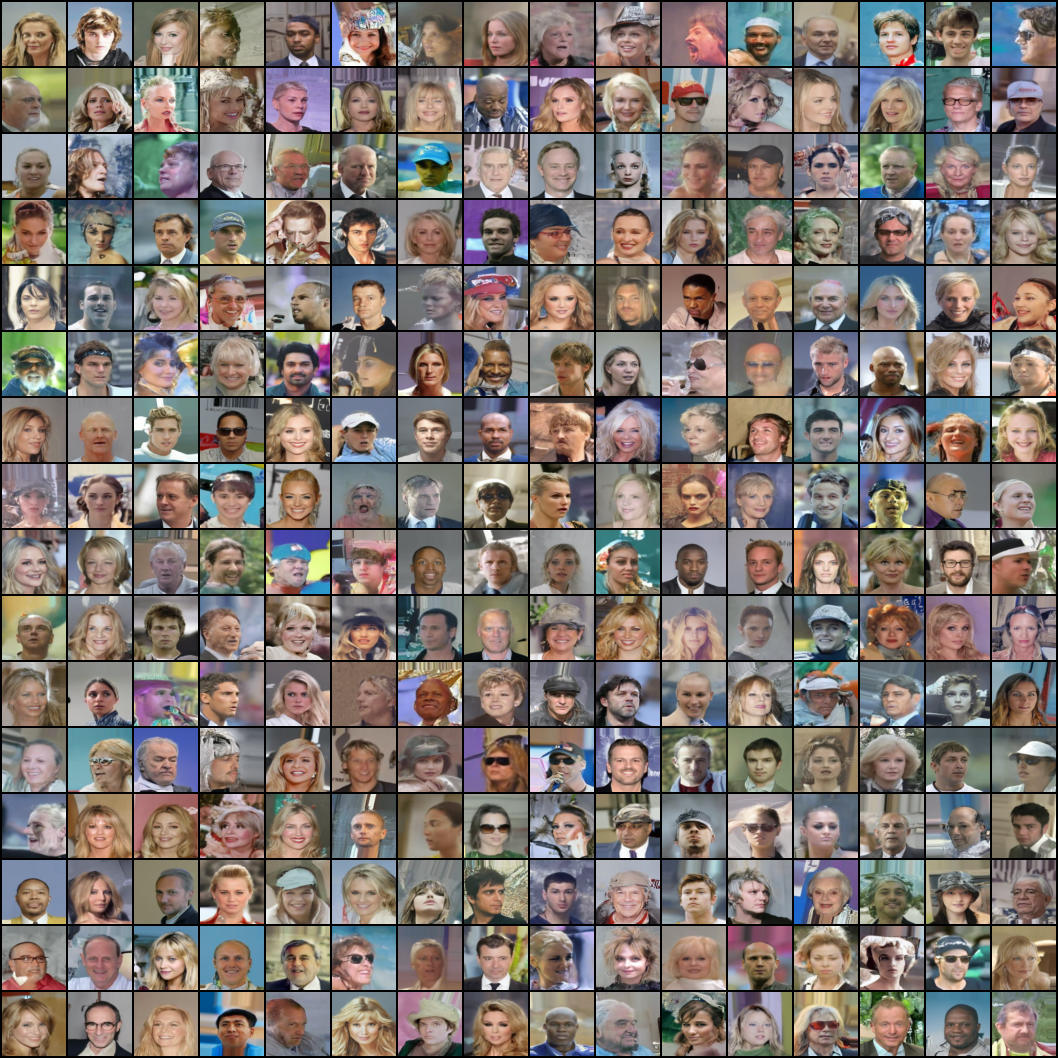}
    \caption{Unconditional CelebA samples}
    \label{fig:celeba_unconditional}
\end{subfigure}%
\begin{subfigure}{.5\textwidth}
    \centering
    \includegraphics[width=\imgwidth\linewidth]{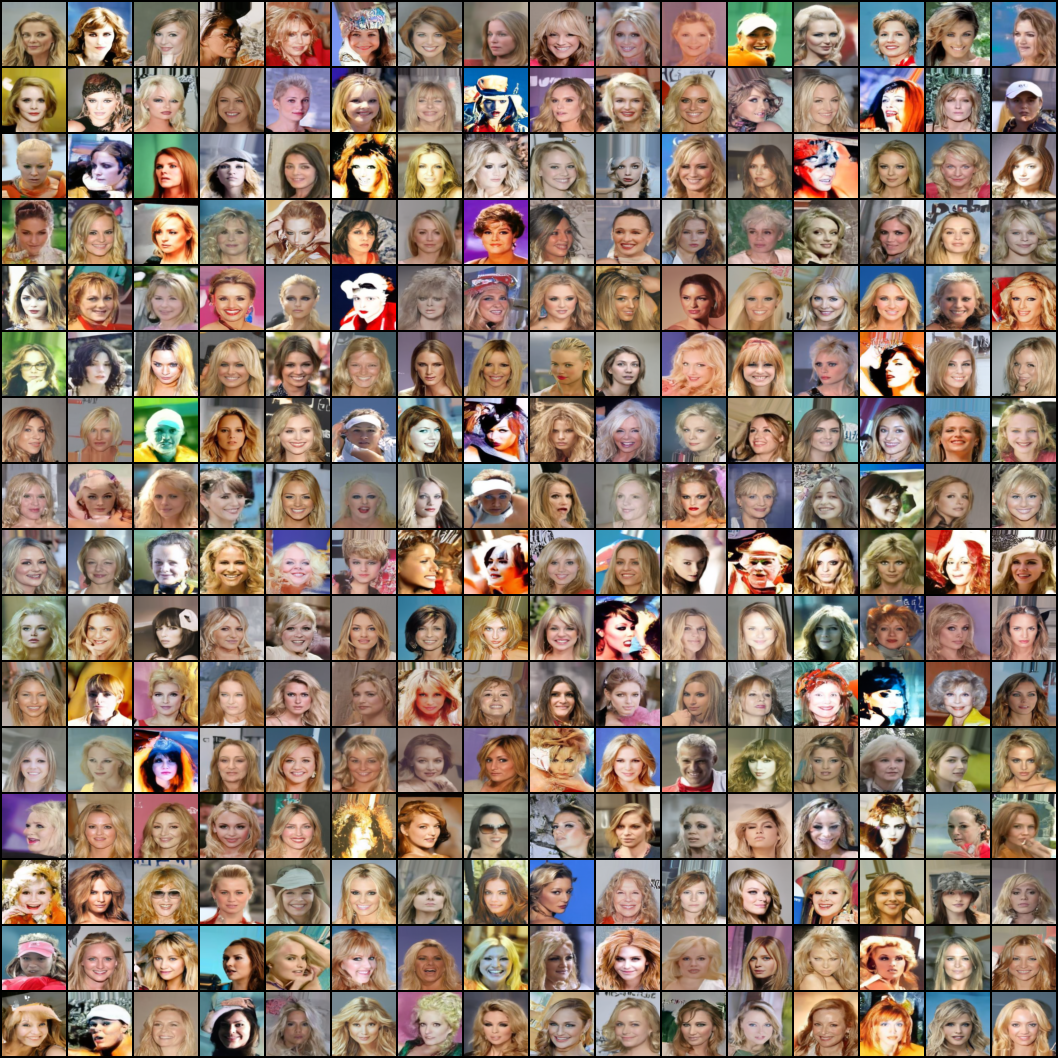}
    \caption{Conditional CelebA samples, s=150}
    \label{fig:celeba_best}
\end{subfigure}%
\\
\begin{subfigure}{.5\textwidth}
    \centering
    \includegraphics[width=\imgwidth\linewidth]{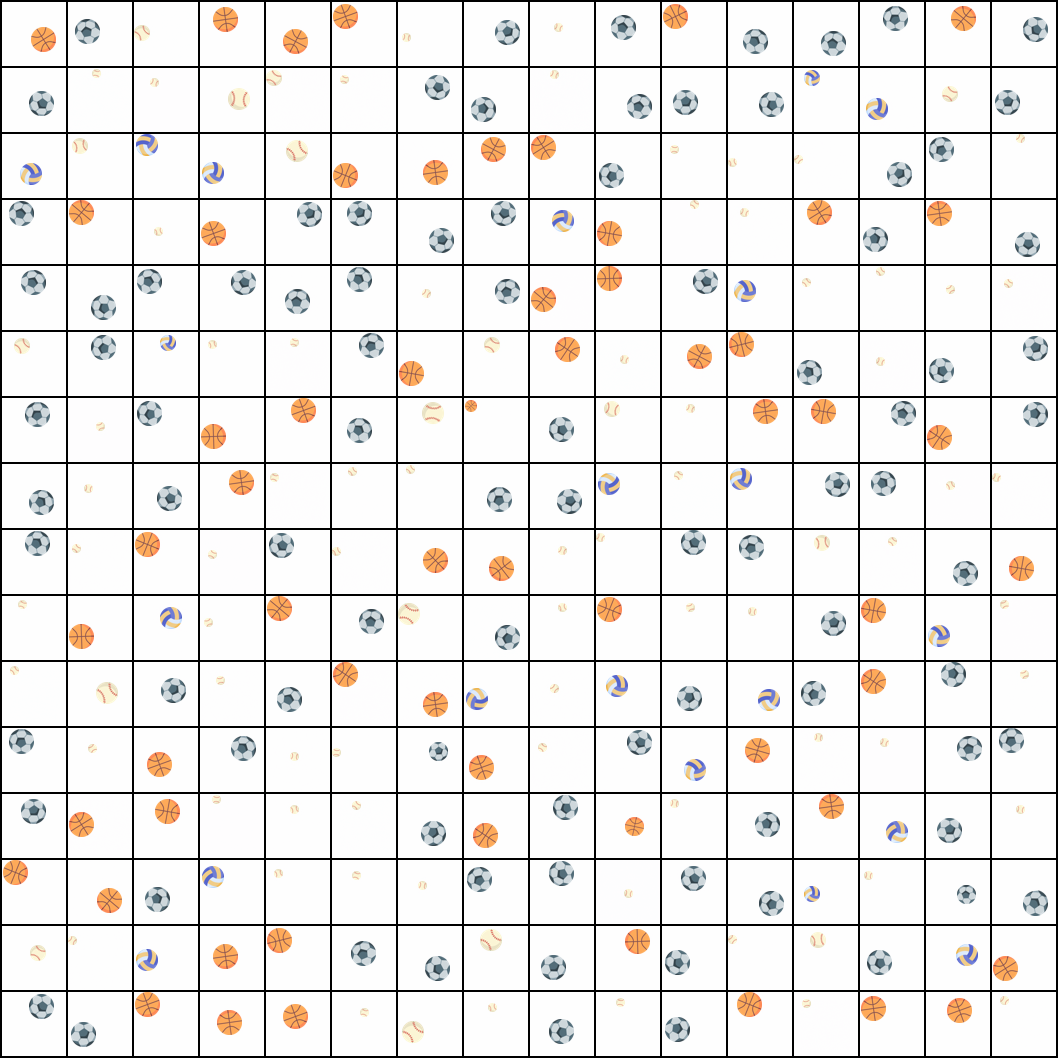}
    \caption{Unconditional SportBalls samples}
    \label{fig:sportballs_unconditional}
\end{subfigure}%
\begin{subfigure}{.5\textwidth}
    \centering
    \includegraphics[width=\imgwidth\linewidth]{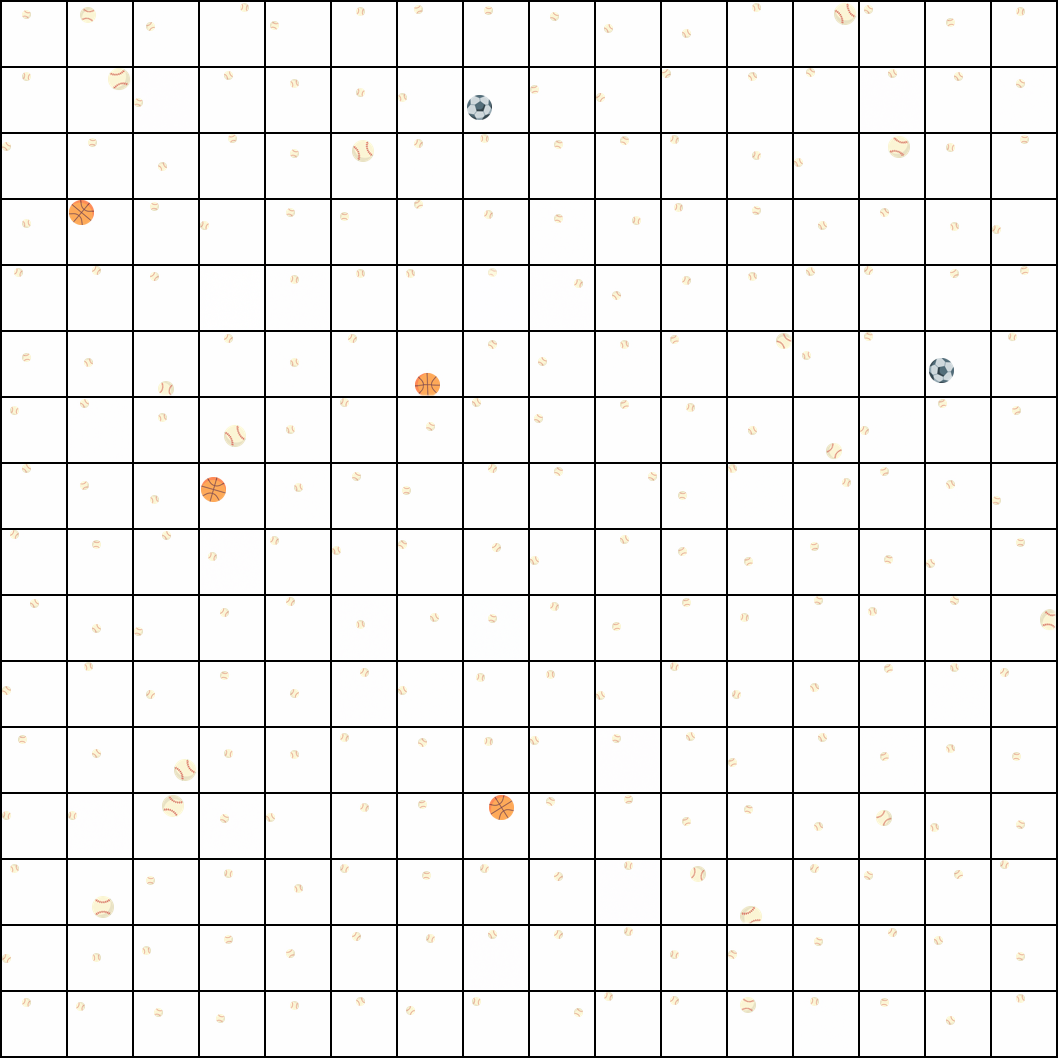}
    \caption{Conditional SportBalls samples, s=15}
    \label{fig:sportballs_best}
\end{subfigure}%
\\
\begin{subfigure}{.5\textwidth}
    \centering
    \includegraphics[width=\imgwidth\linewidth]{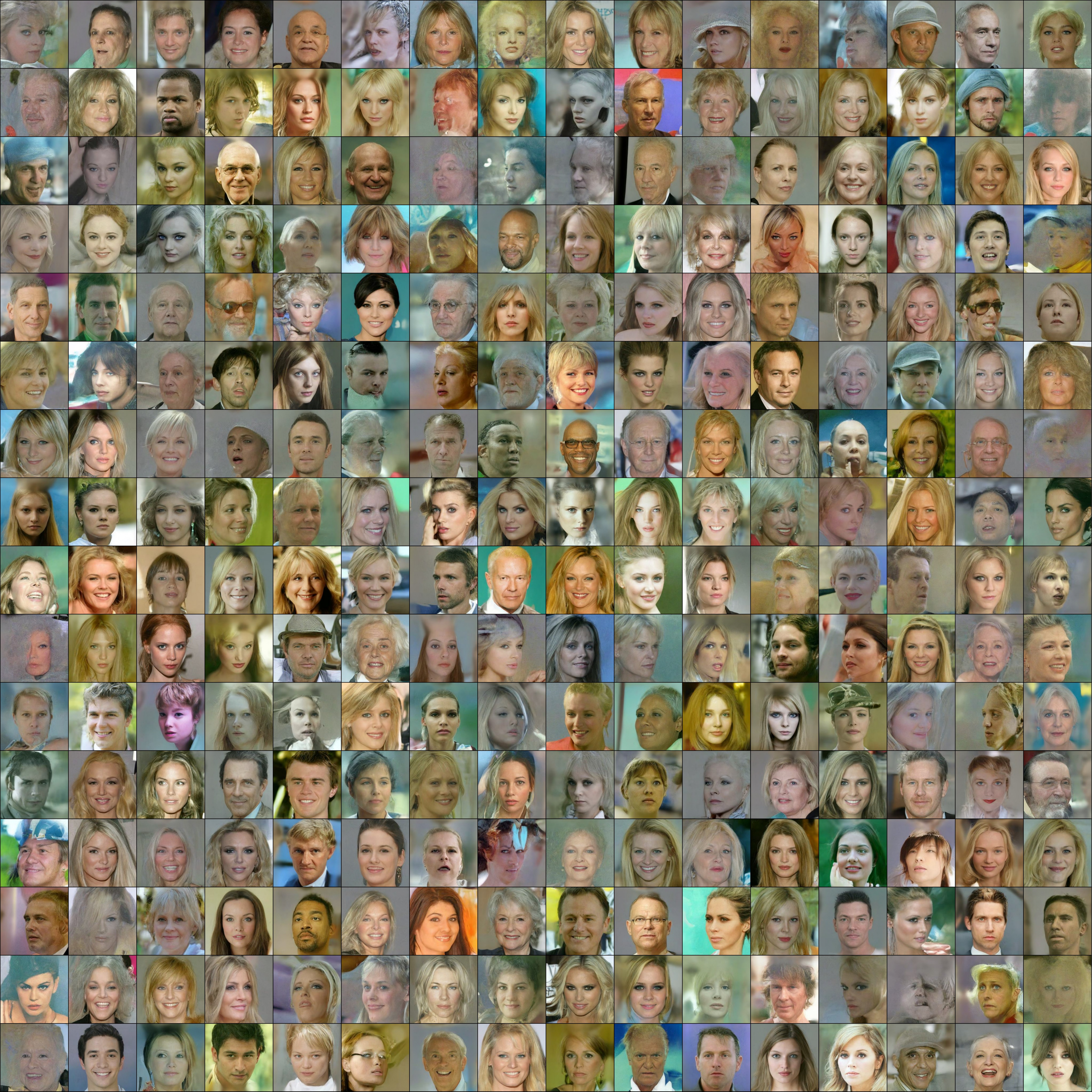}
    \caption{Unconditional CelebA-HQ samples}
    \label{fig:celebahq_unconditional}
\end{subfigure}%
\begin{subfigure}{.5\textwidth}
    \centering
    \includegraphics[width=\imgwidth\linewidth]{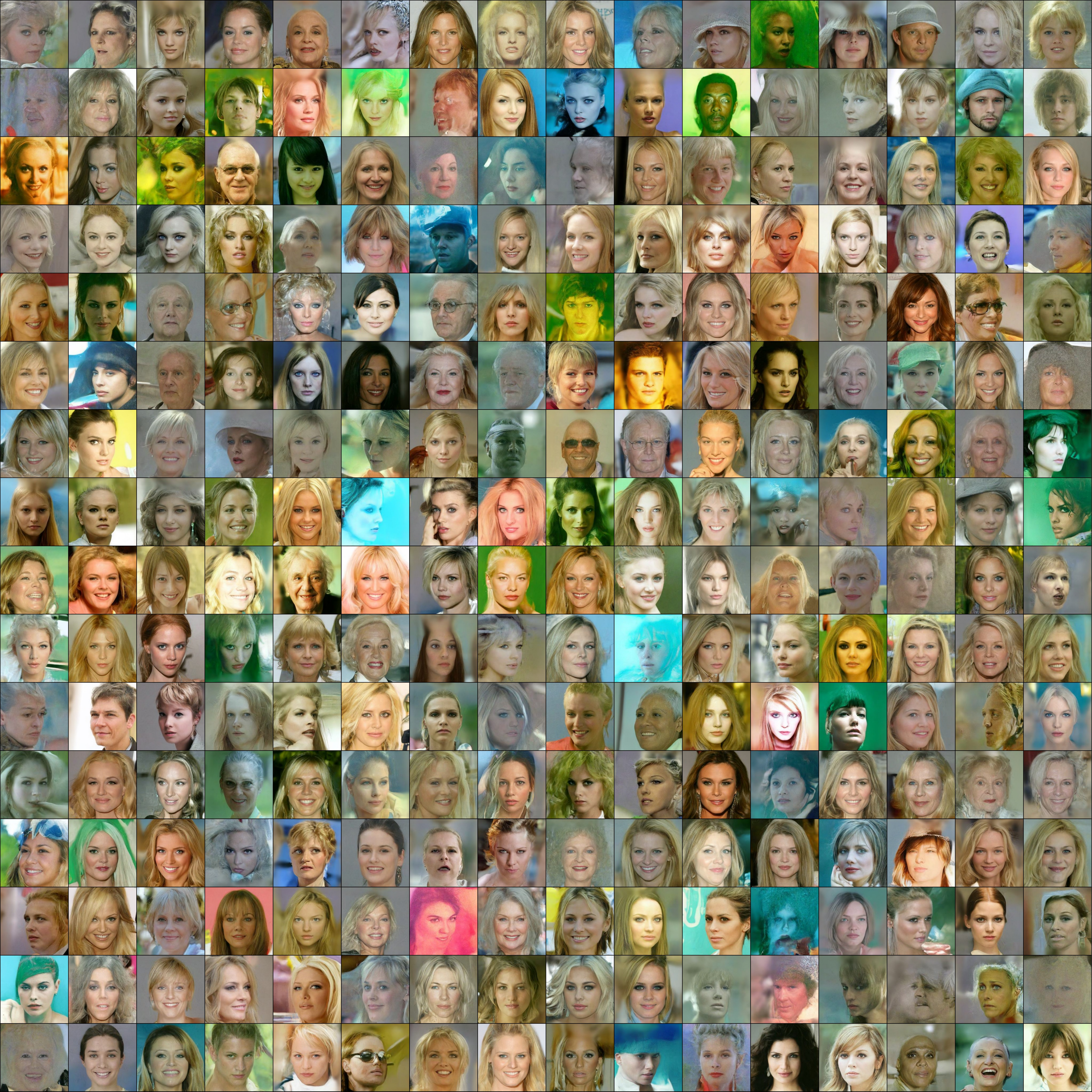}
    \caption{Conditional CelebA-HQ samples, s=10}
    \label{fig:celebahq_best}
\end{subfigure}
\caption{More generations for the different data sets. All conditional samples are generated with our guidance setup of $\xzeroprediction$, 0.99-EMA stabilization and the data set specific guidance scale~$s$. We show \textbf{the first 256} generations.}
\label{fig:appendix_samples}
\end{figure}
\end{document}